\documentclass[a4paper]{article}

\usepackage{INTERSPEECH2019}
\usepackage{multirow}
\usepackage{subcaption}
\usepackage{adjustbox}
\usepackage{chngcntr}
\usepackage{balance}
\usepackage{dblfloatfix}

\title{BERT-DST: Scalable End-to-End Dialogue State Tracking with \\
Bidirectional Encoder Representations from Transformer}
\name{Guan-Lin Chao$^1$, Ian Lane$^2$}
\address{
  Electrical and Computer Engineering$^{1,2}$, Language Technologies Institute$^2$\\
  Carnegie Mellon University, PA, USA}
\email{\{guanlinchao, lane\}@cmu.edu}

\begin{document}

\maketitle
\begin{abstract}
An important yet rarely tackled problem in dialogue state tracking (DST) is scalability for dynamic ontology (\textit{e.g., movie, restaurant}) and unseen slot values. We focus on a specific condition, where the ontology is unknown to the state tracker, but the target slot value (except for \textit{none} and \textit{dontcare}), possibly unseen during training, can be found as word segment in the dialogue context. Prior approaches often rely on candidate generation from n-gram enumeration or slot tagger outputs, which can be inefficient or suffer from error propagation. We propose BERT-DST, an end-to-end dialogue state tracker which directly extracts slot values from the dialogue context. We use BERT as dialogue context encoder whose contextualized language representations are suitable for scalable DST to identify slot values from their semantic context. Furthermore, we employ encoder parameter sharing across all slots with two advantages: (1) Number of parameters does not grow linearly with the ontology. (2) Language representation knowledge can be transferred among slots. Empirical evaluation shows BERT-DST with cross-slot parameter sharing outperforms prior work on the benchmark scalable DST datasets Sim-M and Sim-R, and achieves competitive performance on the standard DSTC2 and WOZ 2.0 datasets.
\end{abstract}
\section{Introduction}
Dialogue state tracking (DST), a core component in today's task-oriented dialogue systems, maintains user's intentional states through the course of a dialogue. The dialogue states predicted by DST are used by the downstream dialogue management component to produce API calls to a backend database and generate responses to the user ~\cite{young2010hidden}. A dialogue state is often expressed as a collection of slot-value pairs.
The set of slots and their possible values are often domain-specific, defined in a \textit{domain ontology}.
Many state-of-the-art approaches operate on a fixed ontology, by performing classification over a predefined set of slot values or iteratively scoring slot-value pairs from the ontology ~\cite{mrkvsic2017nbt, zhong2018glad, ren2018towards}. However, such models can be inefficient or infeasible when the ontology is dynamic (\textit{e.g.}, \textit{movie, restaurant}), innumerable (\textit{e.g.}, \textit{time}), or simply not exposed by an external database~\cite{rastogi2017scalable, xu2018end}.

In this paper, we study this practical problem in DST -- scalability with unknown ontology and unseen slot values, with a specific condition: the target slot value (except for \textit{none} and \textit{dontcare}) always appears as word segment in the dialogue context.
Previous approaches often require a \textit{candidate list}, which can be an exhaustive list of n-grams in the dialogue context or slot tagging outputs from a separate language understanding (LU) module~\cite{rastogi2017scalable, rastogi2018multi, goelflexible}. Using n-gram candidate generation might be inefficient because the number of candidates the DST scorer needs to iterate through is proportional to the length of the dialogue context. Although the LU-generated candidate list can be shorter, the DST scorer cannot recover from missing target candidates incurred by LU errors~\cite{rastogi2017scalable, xu2018end}.

We introduce BERT-DST\footnote{Code is available at https://github.com/guanlinchao/bert-dst}, a scalable end-to-end dialogue state tracker, based on the BERT model~\cite{devlin2018bert}, that directly predicts slot values from the dialogue context with no dependency on candidate generation. In our framework, BERT is adopted to produce contextualized representations of dialogue context (Section~\ref{sec:encoder}), which are used to by the classification and span prediction modules to predict the slot value as \textit{none}, \textit{dontcare} or a text span in the dialogue context (Section~\ref{sec:classification}, ~\ref{sec:span_prediction}). The advantages of using BERT as dialogue context encoder include: (1) The contextualized word representations are suitable for extracting slot values from semantic context. (2) Pre-trained on large-scale language modeling datasets, BERT's word representations are good initialization to be fine-tuned to our DST problem (Section~\ref{sec:bert}).
Moreover, we employ parameter sharing in the BERT dialogue encoder across all slots, which reduces the number of model parameters. Contextualized language representation can also benefit from more training examples of other slots (Section~\ref{sec:parameter_sharing}).
To prevent overfitting, we apply the slot value dropout technique, originally introduced in ~\cite{xu2014targeted, rastogi2017scalable, xu2018end}. This step is critical for extracting unseen slot values from their contextual patterns (Section~\ref{sec:slot_value_dropout}). Through empirical evaluation, we show the effectiveness of BERT-DST with parameter sharing and slot value dropout. BERT-DST achieves state-of-the-art performance of 80.1\% and 89.6\% joint goal accuracy on the benchmark scalable DST datasets Sim-M and Sim-R~\cite{shah2018building}, and competitive results on the standard DSTC2~\cite{henderson2014second} and WOZ 2.0~\cite{wen2017network} datasets (Section~\ref{sec:results}).
\begin{figure*}[t!]
\centering
\includegraphics[width=0.9\textwidth]{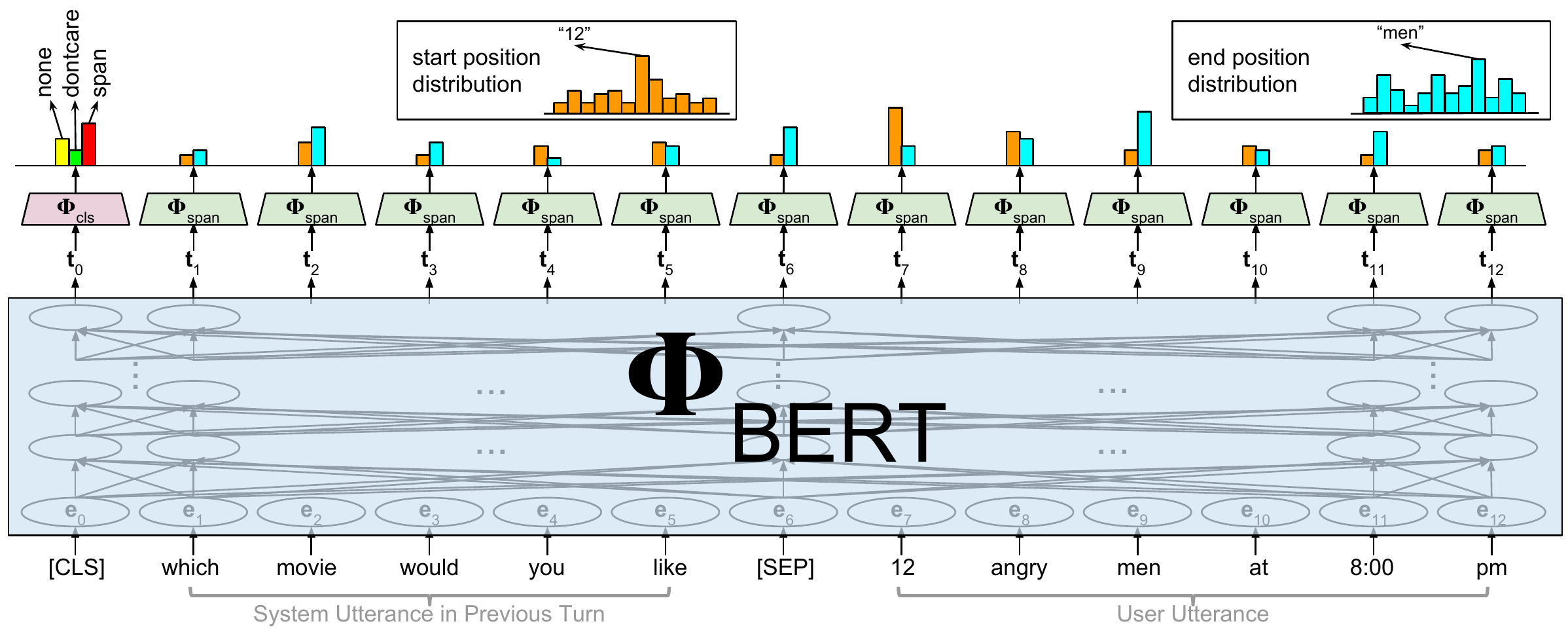}
\caption{Architecture of the proposed BERT-DST framework. The diagram is color-coded such that modules with the same color share the same parameters.
For each user turn, BERT-DST takes as input the recent dialogue context (system utterance in previous turn and the user utterance), and outputs turn-level dialogue state. BERT dialogue context encoding module $\Phi_\text{BERT}$ (blue) produces contextualized sentence-level and token-level representations of the dialogue context. The per-slot classification module $\Phi_\text{cls}$ (red) uses the sentence-level representation to generate a categorical distribution over three types of slot values \{\textit{none}, \textit{dontcare} and \textit{span}\}. The per-slot span prediction module $\Phi_\text{span}$ (green) gathers the token-level representations and output the start and end positions (span) of the slot value. Note that the dialogue context encoding module $\Phi_\text{BERT}$ allows parameter sharing across all slots.
}
\label{fig:bert-dst}
\end{figure*}

\section{BERT}
\label{sec:bert}
In this section, we briefly describe BERT~\cite{devlin2018bert} and how its architecture can be applied to scalable DST in our framework.

BERT is a multi-layer bidirectional Transformer encoder~\cite{vaswani2017attention}, which is a stack of multiple identical layers each containing a multi-head self-attention and a fully-connected sub-layer with residual connections~\cite{he2016deep}. The input to BERT is a sequence of tokens, which can be concatenation of a pair of sentences. The input sequence is prepended by a special \texttt{[CLS]} token whose final hidden state is used as the aggregate sequence representation. The final hidden states of the other tokens are used as token-level representations. Besides word embedding and positional embedding used in the original Transformer model, BERT's input layer adds an additional segment embedding to differentiate tokens from the pair of sentences.

To learn bidirectional contextualized representations and inter-sentence relationship, BERT model is pre-trained on two unsupervised language modeling tasks: masked language modeling~\cite{taylor1953cloze} and next sentence prediction, using the BooksCorpus \cite{zhu2015aligning} and the English Wikipedia corpora. The procedures of language model pre-training are detailed in~\cite{devlin2018bert}. With extra projection layers and fine-tuning the deep structure, BERT has been successfully applied to various tasks such as reading comprehension, named entity recognition, sentiment analysis, \textit{etc}.

Our proposed application of BERT to scalable DST is in spirit similar to the Stanford Question Answering Dataset (SQuAD) task~\cite{rajpurkar2016squad}. In SQuAD, the input is a question and a reading passage. If the reading paragraph contains the answer to the question, the output is a segment of text from the paragraph, represented by its span (start and end positions). Otherwise, the model should output \textit{unanswerable}. Similarly, in our targeted case of scalable DST, a slot's value can be \textit{none}, \textit{dontcare}, or a word segment from the dialogue context. Our proposed framework uses BERT's contextualized sentence-level and token-level representations to determine the type of slot value (\textit{none}, \textit{dontcare}, or \textit{span}), and the span of the specified slot value from the dialogue context. Using BERT as dialogue context encoder provides the following advantages. The contextualized word representations are suitable for extracting slot values from contextual patterns. With large-scale language model pre-training, BERT's word representations are good initialization to be fine-tuned to our DST problem.

\section{BERT-DST}
\label{sec:BERT-DST}
In this section, we describe in detail the proposed BERT-DST framework, as shown in Figure~\ref{fig:bert-dst}.
For each user turn, BERT-DST takes the recent dialogue context as input and outputs the turn-level dialogue state.
First, the dialogue context input is encoded by the BERT-based encoding module to produce contextualized sentence-level and token-level representations.
The sentence-level representation is then used by the classification module to generate a categorical distribution over three types of slot values: \textit{none}, \textit{dontcare} or a \textit{span} from the input. The span prediction module gathers the token-level representations and outputs the slot value's start and end positions. Finally, an update mechanism is used to track dialogue states across turns.

\subsection{Dialogue Context Encoding Module}
\label{sec:encoder}
The dialogue context encoding module is based on BERT.
We use the system utterance from the previous turn and the current turn user utterance as dialogue context input, represented as a token sequence in BERT's input format. The first token is \texttt{[CLS]}, followed by the tokenized system utterance, \texttt{[SEP]}, and tokenized user utterance. Let $[x_0, x_1, \cdots, x_n]$ denote the input token sequence. BERT's input layer embeds each token $x_i$ into an embedding $\mathbf{e}_i$, which is the sum of three embeddings:
\begin{align}
    \mathrm{BERTinput}(x_i) &= E_\text{tok}(x_i) + E_\text{seg}(i) + E_\text{pos}(i) \nonumber \\
    &= \mathbf{e}_i \in \mathbb{R}^d, ~~\forall~0 \leq i \leq n
\end{align}
where $E_\text{tok}(x_i)$ is WordPiece embedding~\cite{wu2016google} for token $x_i$, $E_\text{seg}(i) \in \{\mathbf{e}_\text{first}, \mathbf{e}_\text{second}\}$ is segment embedding whose value is determined by whether the token belongs to the first or second sentence, and $E_\text{pos}(i)$ is positional embedding~\cite{vaswani2018tensor2tensor} for the $i$-th token.

The embedded input sequence $[\mathbf{e}_0, \cdots, \mathbf{e}_n]$ is then passed to BERT's bidirectional Transformer encoder, whose final hidden states are denoted by $[\mathbf{t}_0, \cdots, \mathbf{t}_n]$.
\begin{align}
    & [\mathbf{t}_0, \cdots, \mathbf{t}_n] = \mathrm{BiTransformer}([\mathbf{e}_0, \cdots, \mathbf{e}_n]) \nonumber \\
    & \mathbf{t}_i \in \mathbb{R}^d, ~~\forall~0 \leq i \leq n
\end{align}
The contextualized sentence-level representation $\mathbf{t}_0$, \textit{i.e.}, the final state corresponding to the \texttt{[CLS]} token, is passed to the classification module. The contextualized token-level representations $[\mathbf{t}_1, \cdots, \mathbf{t}_n]$ are used by the span prediction module.

The parameters in the dialogue context encoding module, denoted by $\Phi_\text{BERT}$, are initialized from a pre-trained BERT checkpoint and then fine-tuned on our DST dataset.

\subsection{Classification Module}
\label{sec:classification}
The classification module's input is the sentence-level representation $\mathbf{t}_0$ from the dialogue context encoding module. For each slot $s \in S$ in the collection of all informable slots $S$, the classification module predicts the value of $s$ to be one of the three classes \{\textit{none}, \textit{dontcare}, \textit{span}\} using linear projection and softmax.
\begin{align}
    & \mathbf{a}^s = \mathbf{W}^s_{cls} \mathbf{t}_0 + \mathbf{b}_\text{cls}^s = [a^s_\text{none}, a^s_\text{dontcare}, a^s_\text{span}] \in \mathbb{R}^3 \\
    & \mathbf{p}^s = \mathrm{softmax}(\mathbf{a}^s) = [p^s_\text{none}, p^s_\text{dontcare}, p^s_\text{span}] \\
    & \mathrm{slot\_value}^s = \mathrm{argmax}_{c \in \{\text{none, dontcare, span}\}} (p^s_c)
\end{align}
The per-slot classification parameters, denoted by $\Phi^s_\text{cls} = \{\mathbf{W}^s_\text{cls}, \mathbf{b}_\text{cls}\}$, are trained from scratch on our DST dataset.

\subsection{Span Prediction Module}
\label{sec:span_prediction}
For each slot $s \in S$, the span prediction module takes as input the token-level representations $[\mathbf{t}_1, \cdots, \mathbf{t}_n]$ from the dialogue context encoding module.
Each token representation $\mathbf{t}_i$ is linearly projected through a common layer whose output values $\alpha^s_i$ and $\beta^s_i$ correspond to start and end positions respectively. Softmax is then applied to the position values to produce a probability distribution over all tokens, by which the slot value span (start and end positions) of $s$ can be determined.
\begin{align}
    [\alpha^s_i, \beta^s_i] &= \mathbf{W}_\text{span}^s \mathbf{t}_i + \mathbf{b}^s_\text{span} \in \mathbb{R}^2, \forall~1 \leq i \leq n \\
    \mathbf{p}^s_\alpha &= \mathrm{softmax}(\mathbf{\alpha}^s) \\
    \mathbf{p}^s_\beta &= \mathrm{softmax}(\mathbf{\beta}^s) \\
    \mathrm{start\_pos}^s &= \mathrm{argmax}_i(p^s_{\alpha, i}) \\
    \mathrm{end\_pos}^s &= \mathrm{argmax}_i(p^s_{\beta, i})
\end{align}
The per-slot span prediction parameters, denoted by $\Phi^s_\text{span} = \{\mathbf{W}^s_\text{span}, \mathbf{b}^s_\text{span}\}$, are trained from scratch on our DST dataset.

\subsection{Dialogue State Update Mechanism}
To track dialogue states across turns, we employ a rule-based update mechanism. In each turn, if the model's turn prediction for a slot is \textit{dontcare} or a specified value (\textit{i.e.}, any value other than \textit{none}), it will be used to update the dialogue state. Otherwise, the dialogue state of the slot remains the same as the previous turn.

\subsection{Parameter Sharing}
\label{sec:parameter_sharing}
Although our classification and span prediction modules are slot-specific, we notice that the contextualized representations generated by the dialogue context encoding module can be shared among slots; \textit{i.e.}, we can apply parameter sharing in the dialogue context encoding module across all slots. Sharing dialogue context encoder parameters $\Phi_\text{BERT}$ across all slots not only drastically reduces the number of model parameters. It also allows knowledge transfer among slots, which may potentially benefit contextual relation understanding. In the following sections, we call the joint architecture of slot-specific BERT-DST models as \textbf{BERT-DST\_SS} and the BERT-DST model with encoding module parameter sharing as \textbf{BERT-DST\_PS}.

\subsection{Slot Value Dropout}
\label{sec:slot_value_dropout}
Slot value dropout, or targeted feature dropout, was originally proposed to address the under-training problem of contextual features in slot-filling~\cite{xu2014targeted}.
The problem happens when models tend to overfit to frequent slot values in training data instead of learning contextual patterns, which adversely harms the performance on OOV slot values.
To improve the robustness 
for unseen slot values, in the training phase, we replace each of the target slot value tokens by a special \texttt{[UNK]} token at a certain probability.
\section{Experiments}
We evaluate our models using \textit{joint goal accuracy}~\cite{henderson2014second}, a standard metric for DST. The model's prediction has to jointly match all the informable slot labels to be considered correct.

\subsection{Datasets}
We evaluate our models on four benchmark datasets: Sim-M, Sim-R~\cite{shah2018building}, DSTC2~\cite{henderson2014second} and WOZ 2.0~\cite{wen2017network}.
The statistics of the datasets are shown in Table~\ref{tab:datasets} in Appendix.

Sim-M and Sim-R are specialized for scalable DST, which contain human-paraphrased simulated dialogues in the movie and restaurant domains.
They have span annotations for all specified slot values (\textit{i.e.} values other than \textit{none} and \textit{dontcare}) in the system and user utterances.
In the event that the target slot value has multiple spans in the dialogue context, we use the span of the last occurrence as reference.
The prevalence of out-of-vocabulary (OOV) values in Sim-M's \textit{movie} and Sim-R's \textit{restaurant\_name} slots makes them particularly challenging and suitable for scalable DST evaluation.

DSTC2 and WOZ 2.0 are standard benchmarks for task-oriented dialogue systems, which are both in the restaurant domain and share the same ontology.
In DSTC2, automatic speech recognition (ASR) hypotheses of user utterances are provided to assess DST models' robustness against ASR errors, so we use the top ASR hypothesis for validation and testing. In WOZ 2.0, the user interface is typing and collection of user utterances exhibit higher degree of lexical variation. ASR errors and flexible language use can cause problems in defining slot value spans, which is basis for our targeted scalable DST condition. The problem arises when erroneous ASR hypotheses do not contain a user's intended specified value or the user uses a \textit{creative} expression in which a clear boundary of a value's span can be hard to define. For example, "My wife thinks she likes international but I don't want to take out a loan." is annotated with \texttt{price\_range=cheap}. Such challenging instances in DSTC2 and WOZ 2.0 set the performance upper bound for our proposed scalable DST framework.

Note that in the evaluation, we do not apply an output canonicalization step to handle other possible valid variations of span. Therefore our model's predicted slot value span has to exactly match the label span to be considered correct.

\subsection{Implementation Details}
We use the pre-trained \textit{[BERT-Base, Uncased]} model which has 12 hidden layers of 768 units and 12 self-attention heads for lower-cased input text\footnote{https://github.com/google-research/bert}. The span prediction loss for \textit{\{none, dontcar\}} slots is set to zero. The total loss is defined as  $(0.8\mathcal{L}^\text{xent}_\text{cls} + 0.1\mathcal{L}^\text{xent}_\text{span\_start} + 0.1 \mathcal{L}^\text{xent}_\text{span\_end})$, where $\mathcal{L}^\text{xent}$ denotes the cross entropy loss for the corresponding prediction target.
We update all layers in the model using ADAM optimization~\cite{kingma2014adam} with an initial learning rate $2e^{-5}$ and early stopping on the validation set. During training, we use 30\% dropout rate~\cite{srivastava2014dropout} on the dialogue context encoder outputs. Various rates of slot value dropout are experimented and reported in Section~\ref{sec:results}.
\section{Results and Discussion}
\label{sec:results}

\begin{table}
\centering
 \begin{tabular}{l  l  l} 
 \toprule
 DST Models & Sim-M & Sim-R \\
 \midrule
 DST + LU Candidates~\cite{rastogi2018multi} & 50.4\% & 87.1\% \\
 DST + Oracle Candidates$^\dagger$~\cite{rastogi2017scalable} & 96.8\% & 94.4\% \\
 \midrule
 BERT-DST\_SS & 71.6\% & 87.4\% \\
 ~~~~+ slot value dropout & 76.3\%* & 87.6\% \\
 BERT-DST\_PS & 72.3\% & 88.6\%* \\
 ~~~~+ slot value dropout & \textbf{80.1\%}* & \textbf{89.6\%}* \\
 \bottomrule
\end{tabular}
\caption{Comparison with prior approaches on Sim-M and Sim-R datasets (joint goal accuracy).
* indicates statistically significant improvement over BERT-DST model (paired sample t-test; $p < 0.01$). 
$^\dagger$ indicates the corresponding model should be considered as a kind of oracle because the candidates are ground truth slot-tagging labels, \textit{i.e.} the targeted slot value is guaranteed to be in the candidate list and considered by DST.}
\label{tab:results_sim}
\end{table}

\begin{figure}[t]
\centering
\vspace*{-0.4cm}
\includegraphics[width=\linewidth]{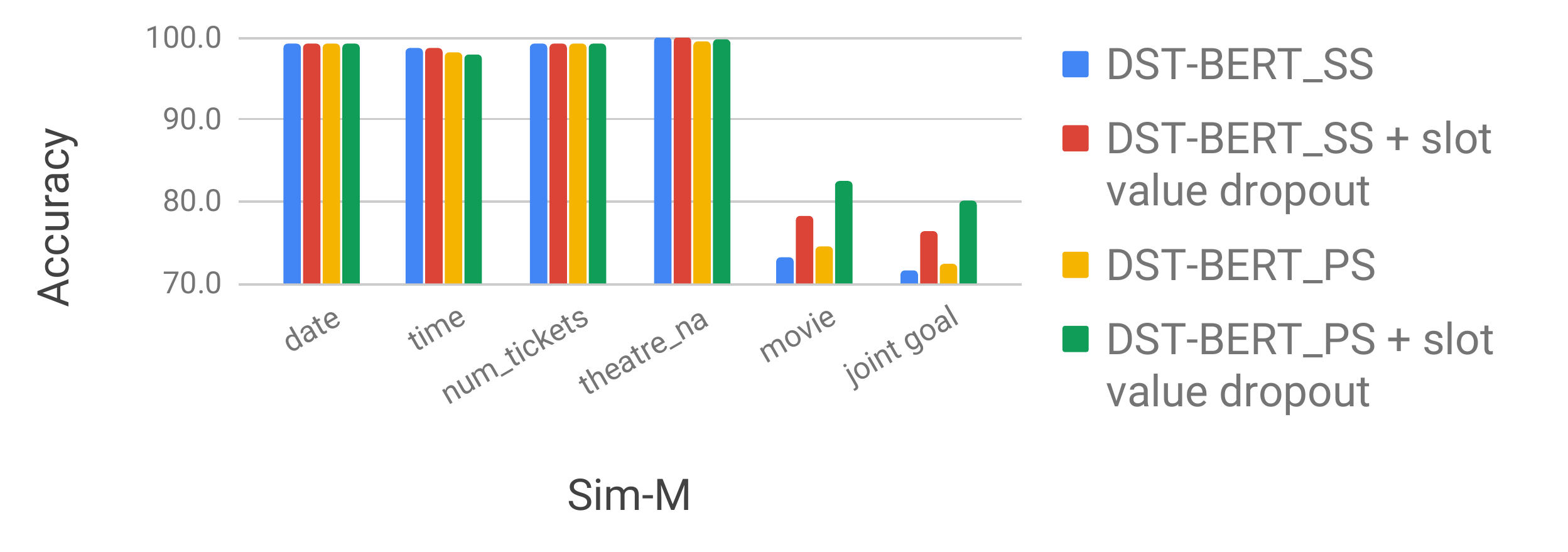}\\
\vspace*{-0.2cm}
\includegraphics[width=\linewidth]{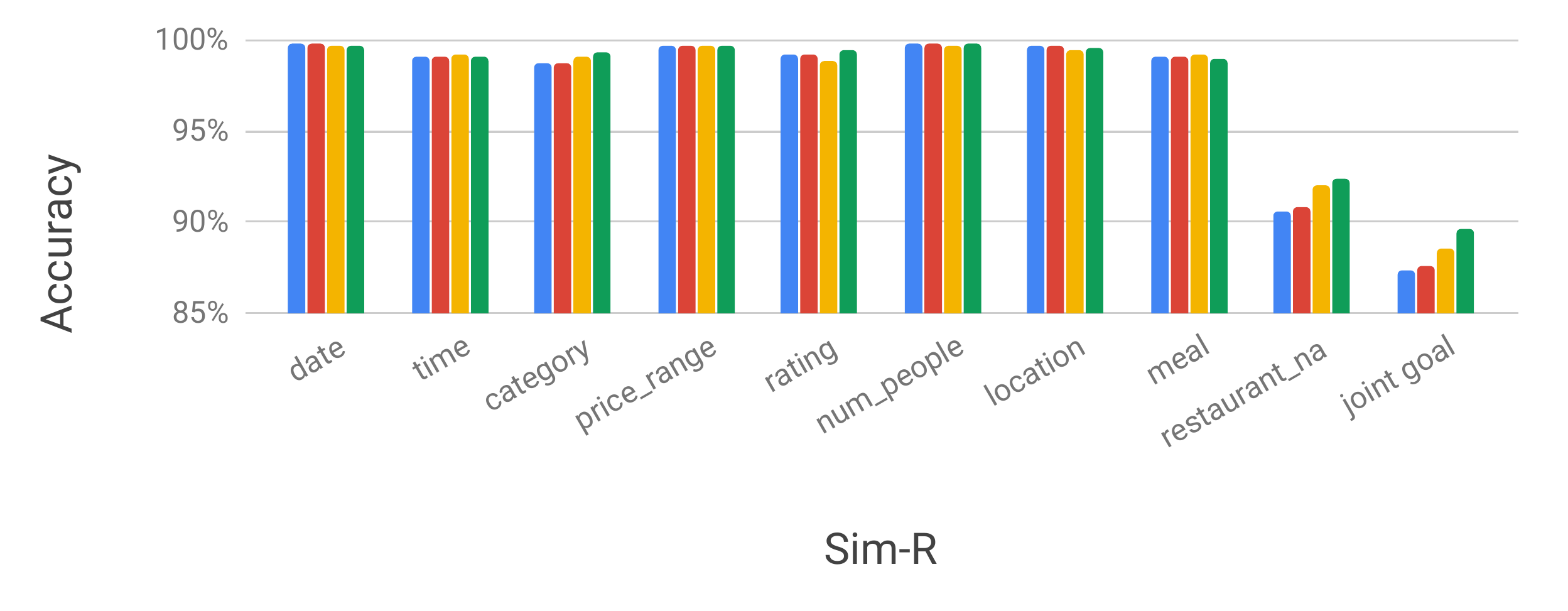}
\vspace*{-0.8cm}
\caption{Per-slot and joint goal accuracy of the proposed models on Sim-M and Sim-R datasets.
}
\label{fig:breakdown}
\end{figure}

Table~\ref{tab:results_sim} presents the performance of the proposed BERT-DST models compared to prior work on the scalable DST datasets Sim-M and Sim-R.
In \cite{rastogi2018multi, rastogi2017scalable}, the DST component scores slot values from a candidate list, which is slot tagging predictions of a jointly-trained language understanding component (\textit{DST + LU Candidates}), or the ground truth slot tagging labels (\textit{DST + Oracle Candidates}).
We compare our proposed model with the (\textit{DST + LU Candidates}) baseline because in practice an oracle candidate list that always contains the target slot label is rarely available.
On both Sim-M and Sim-R, BERT-DST\_SS outperforms the baseline model.
We attribute the performance gain to the effective contextualized representations obtained from the BERT dialogue encoding module.
BERT-DST\_PS with slot value dropout achieves further statistically significant improvement over BERT-DST\_SS. 
The comparison of per-slot and joint goal accuracy of the different BERT-DST models is shown in Figure~\ref{fig:breakdown}. We observe that it is mainly the slots with OOV values (\textit{movie} for Sim-M and \textit{restaurant\_name} for Sim-R) that benefit from the encoder parameter sharing and slot value dropout techniques. The accuracy improvement on these bottleneck slots eventually leads to gain in the joint goal accuracy.

\begin{figure}[t]
\centering
\vspace*{-0.4cm}
\includegraphics[width=\linewidth]{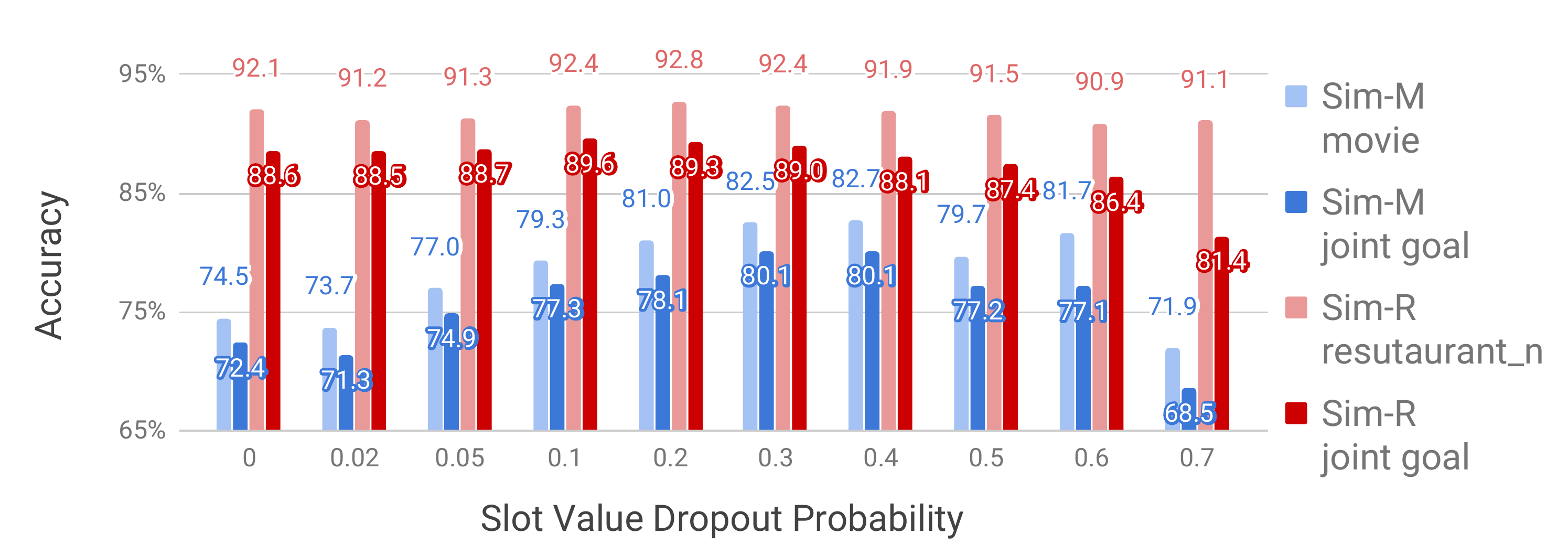}
\vspace*{-0.5cm}
\caption{Comparison of different slot value dropout probabilities of the BERT-DST\_PS model on Sim-M and Sim-R datasets.}
\label{fig:slot_value_dropout}
\end{figure}

To investigate the effect of slot value dropout, we compare the performance with different slot value dropout probabilities of BERT-DST\_PS on Sim-M and Sim-R datasets, as shown in Figure~\ref{fig:slot_value_dropout}. While a proper selection of slot value dropout rate can result in slight improvement on Sim-R, the effect of slot value dropout is more pronounced on Sim-M. Because of the high OOV value rate of the \textit{movie} slot (100\% OOV in test set), higher slot value dropout rate can be helpful for extracting unseen slot values from contextual patterns.

\begin{table}[t]
\centering
\begin{tabular}{l  l  l} 
 \toprule
 DST Models & DSTC2 & WOZ 2.0 \\
 \midrule
 DST + LU Candidates~\cite{rastogi2018multi} & 67.0\% & - \\
 DST + n-gram Candidates~\cite{goelflexible} & 68.2$\pm$1.8\% & - \\
 DST + Oracle Candidates~\cite{rastogi2017scalable} & 70.3\% & - \\
 Pointer Network~\cite{xu2018end} & 72.1\% & -\\
 \midrule
 Delex.-Based Model~\cite{mrkvsic2017nbt} & 69.1\% & 70.8\% \\
 Delex. + Semantic Dict.~\cite{mrkvsic2017nbt} & 72.9\% & 83.7\% \\
 Neural Belief Tracker~\cite{mrkvsic2017nbt} & 73.4\% & 84.2\% \\
 GLAD~\cite{zhong2018glad} & 74.5$\pm$0.2\% & 88.1$\pm$0.4\% \\
 StateNet~\cite{ren2018towards} & \textbf{75.5\%} & \textbf{88.9\%} \\
 \midrule
 BERT-DST\_PS & 69.3$\pm$0.4\% & 87.7$\pm$1.1\% \\
 \bottomrule
\end{tabular}
\caption{
Comparison with prior approaches on DSTC2 and WOZ 2.0 datasets (joint goal accuracy).
We report the average and standard deviation of test set accuracy of 5 model runs with random training data shuffling and normal initialization on classification and span prediction weights.}
\label{tab:results_dst}
\end{table}

Table~\ref{tab:results_dst} presents the performance of BERT-DST with prior approaches on the standard DSTC2 and WOZ 2.0 datasets. Our work is more comparable with the top group frameworks, which are also designed for scalable DST to handle unknown ontology. The middle group of models require a predefined ontology to perform classification or scoring over a predefined set of possible slot values.
On DSTC2, BERT-DST\_PS shows comparable performance with prior scalable DST models, although not as high as the state-of-the-art models.
On WOZ 2.0, BERT-DST\_PS achieves competitive results with state-of-the-art models, which demonstrates BERT-DST's capability in understanding sophisticated language.
Note that it is not our goal to achieve state-of-the-art performance on the standard datasets. Instead, BERT-DST is tasked to handle unknown ontology and unseen slot values and does not require a separate candidate generation module.
\section{Conclusions}
We introduce BERT-DST, a scalable end-to-end dialogue state tracker to handle unknown ontology and unseen slot values. Not requiring candidate value generation, BERT-DST directly predicts slot values from the dialogue context. The key component is the BERT dialogue context encoding module which produces contextualized representations effective for extracting slot values from the contextual patterns. Empirical evaluation on Sim-M and Sim-R datasets shows the efficacy of the proposed BERT-DST model with the slot value dropout technique and encoder parameter sharing across all slots.
\balance
\begin{figure*}[b!]
\centering
\includegraphics[width=0.85\linewidth]{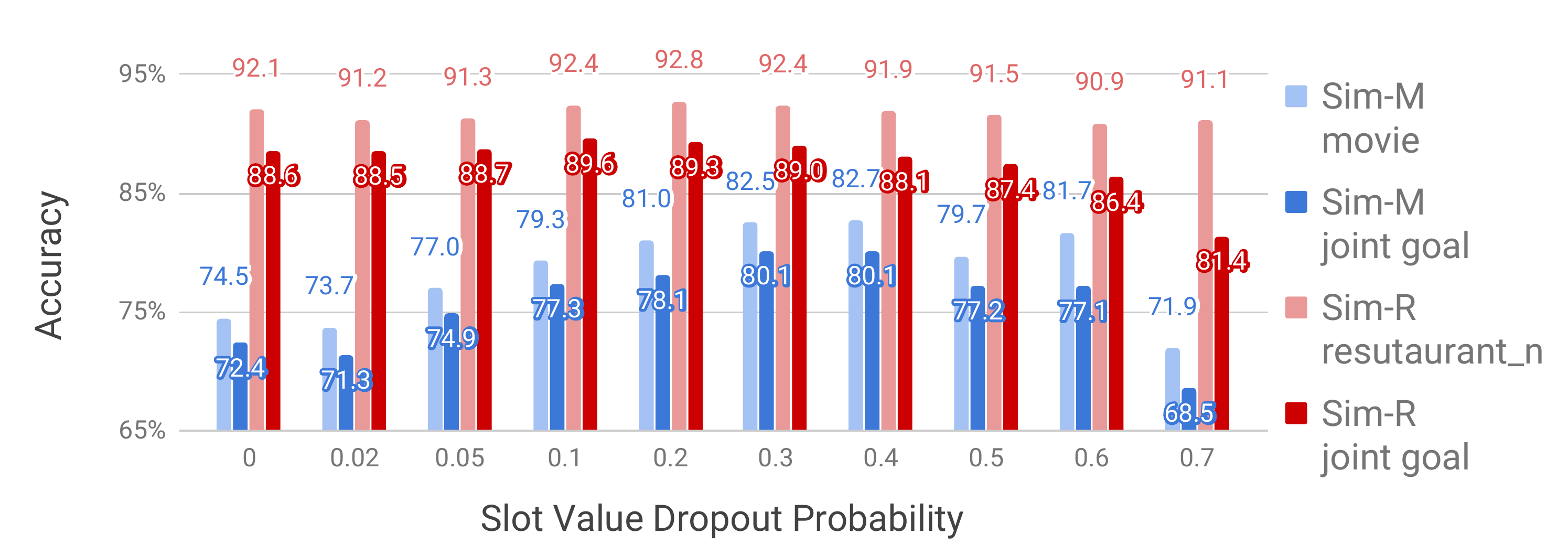}
\caption{(Larger version of Figure~\ref{fig:slot_value_dropout}) Comparison of different slot value dropout probabilities of the BERT-DST\_PS model on Sim-M and Sim-R datasets.}
\vspace*{-2.5cm}
\end{figure*}

\bibliographystyle{IEEEtran}
\bibliography{guanlinc}
\appendix

\section{Appendix}

\begin{table}[!h]
\centering
\begin{adjustbox}{max width=\linewidth}
 \begin{tabular}{l | l | l} 
 \toprule
 \multirow{2}{*}{Datasets} & \# Dialogues & \multirow{2}{*}{Slots}\\
 & (train, dev, test)\\
 \midrule
 \multirow{3}{*}{Sim-M} & \multirow{3}{*}{384, 120, 264} & date, time, num\_tickets,\\
 & & theatre\_name,\\
 & &\textbf{movie} (5/5; 26/26)\\
 \hline
 \multirow{5}{*}{Sim-R} & \multirow{5}{*}{1116, 349, 775} & date, time, category, \\
 & & price\_range, rating, num\_people, \\
 & & location, meal, \\
 & & \textbf{restaurant\_name} (5/19; 9/23)\\
 \hline
 \multirow{2}{*}{DSTC2} & \multirow{2}{*}{1612, 506, 1117} & area, price range,\\
 & & \textbf{food} (1/73; 0/74)\\
 \hline
 \multirow{2}{*}{WOZ 2.0} & \multirow{2}{*}{600, 200, 400} & \textbf{area} (0/6; 1/7), price range,\\
 & & \textbf{food} (1/65; 2/72)\\
 \bottomrule
\end{tabular}
\end{adjustbox}
\caption{Dataset statistics. The number of dialogues is given for train, dev and test sets respectively. The slots containing OOV values are marked in bold. Parentheses represent \\
(\# unique OOV values in dev set / \# unique values in dev set; \\
\# unique OOV values in test set / \# unique values in test set).
}
\label{tab:datasets}
\end{table}

\end{document}